\documentclass[conference]{IEEEtran}
\IEEEoverridecommandlockouts
\usepackage{cite}
\usepackage{amsmath,amssymb,amsfonts}
\usepackage{algorithmic}
\usepackage{graphicx}
\usepackage{textcomp}
\usepackage{xcolor}
\usepackage{comment}
\usepackage{ulem}
\usepackage{multirow}

\def\BibTeX{{\rm B\kern-.05em{\sc i\kern-.025em b}\kern-.08em
    T\kern-.1667em\lower.7ex\hbox{E}\kern-.125emX}}

\colorlet{FigColor}{green!30!blue!70!black!100!}
\colorlet{darkBlue}{blue!30!black!70!}

\begin{document}

\title{Actuator Trajectory Planning for UAVs with Overhead Manipulator using Reinforcement Learning}


\author{\IEEEauthorblockN{Hazim Alzorgan\IEEEauthorrefmark{1}, Abolfazl Razi\IEEEauthorrefmark{1}, Ata Jahangir Moshayedi\IEEEauthorrefmark{2}} \
\IEEEauthorblockA{\IEEEauthorrefmark{1}\textit{School of Computer Science, Clemson University, Clemson, SC}\
Emails: \{halzorg,arazi\}@clemson.edu} \
\IEEEauthorblockA{\IEEEauthorrefmark{2}\textit{School of Information Engineering, Jiangxi University of Sci. \& Tech., Jiangxi, China}\
Email: ajm@jxust.edu.cn}}

\maketitle

\begin{abstract}


In this paper, we investigate the operation of an aerial manipulator system, namely an Unmanned Aerial Vehicle (UAV) equipped with a controllable arm with two degrees of freedom to carry out actuation tasks on the fly. Our solution is based on employing a Q-learning method to control the trajectory of the tip of the arm, also called \textit{end-effector}. More specifically, we develop a motion planning model based on Time To Collision (TTC), which enables a quadrotor UAV to navigate around obstacles while ensuring the manipulator's reachability. Additionally, we utilize a model-based Q-learning model to independently track and control the desired trajectory of the manipulator's end-effector, given an arbitrary baseline trajectory for the UAV platform. 
Such a combination enables a variety of actuation tasks such as high-altitude welding, structural monitoring and repair, battery replacement, gutter cleaning, sky scrapper cleaning, and power line maintenance in hard-to-reach and risky environments while retaining compatibility with flight control firmware. Our RL-based control mechanism results in a robust control strategy that can handle uncertainties in the motion of the UAV, offering promising performance. Specifically, our method achieves 92\% accuracy in terms of average displacement error (i.e. the mean distance between the target and obtained trajectory points) using Q-learning with 15,000 episodes \footnote{This material is based upon the work supported by the National Science Foundation under Grant Numbers 2204721 and 2204445.}.
\end{abstract}


\section{Introduction}
Unmanned Aerial Vehicles (UAVs) are being heavily utilized 
in numerous applications for their high levels of agility and flexible maneuverability compared to Unmanned Ground Vehicles (UGV) and legged robotics. They are involved in a wide range of perception-based applications, including surveillance, landscaping, smart agriculture, structural monitoring, search and rescue, and border control to  name only a few \cite{b1,b2,b3}. Perhaps among the most popular forms of such vehicles are the quad-rotor drones, equipped with four propellers, that offer high maneuverability, stable flight, and large payload capacity with respect to their size.

In recent years, there has been a gradual paradigm shift in the use of drones. Alongside their traditional role in monitoring tasks, aerial manipulators have been receiving increasing attention. The core idea is to use hybrid systems taking advantage of the agility of quad-rotors for flexible mobility as well as the manipulators' capability to carry out complicated tasks on the fly using AI-based control. Such integration opens the door to a wider range of applications beyond the conventional passive perception applications of UAVs, moving towards active functionalities, particularly in complex, hazardous, and hard-to-reach environments.

Recently, several works have emerged showcasing such active controllers for aerial manipulators. For instance, in \cite{UAV_docking}, a three-armed manipulator was used to perform landing and docking tasks in uneven terrain and high-altitude situations. Similarly, \cite{Aerial_manip} proposed an image-based impedance force controller for accurate force tracking of an aerial manipulator.

A recent review by \cite{UAV_Rev} shows that the majority of control strategies have primarily relied on physics-based approaches with offline data processing, lacking the robustness to operate in highly dynamic environments with moving obstacles \cite{Future of UAV}. 
In this paper, we aim to explore the potential of dynamic path planning by developing a simulation environment that emulates the behavior of a hybrid system, where the trajectory of the arm-tip (also called end-effector in this paper) depends on the main platform trajectory as well as the arm manipulators' control signal. We train a Q-learning-based controller using a dataset of state-action pairs collected from the simulation environment while implementing a Time to Collision (TTC)-based motion planning method to design an obstacle-avoiding trajectory for the UAV, demonstrated by drawing an arbitrary end-effector target trajectory. 

\begin{figure}[htbp]
\centerline{\includegraphics[height=1.8 in, width = 3 in]{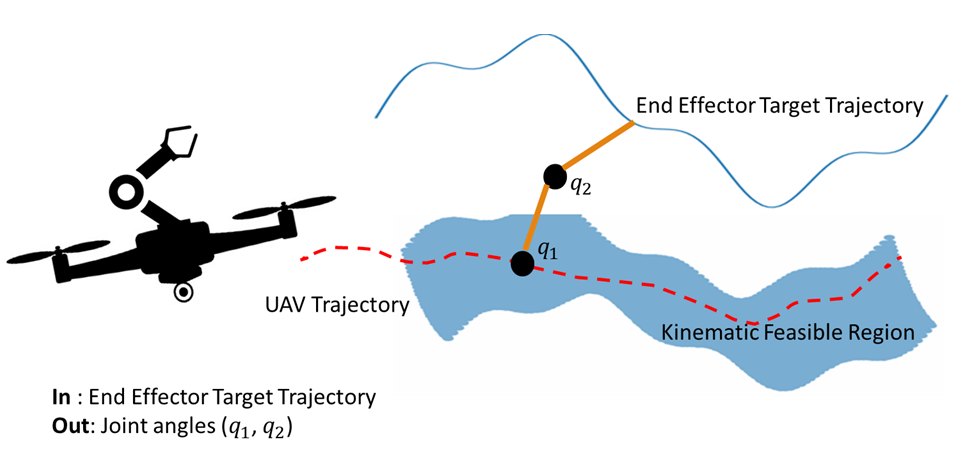}}
\caption{Left: Drone equipped with a manipulator, Right: Demonstration of the base trajectory (solid-blue line) and end-effector trajectory (dashed-red line). The approximate feasible region for the base trajectory is shown by the shaded-blue region.}
\label{fig:sysmodel}
\vspace{-0.2 in}
\end{figure}

Our proposed approach shown in Fig. \ref{fig:sysmodel} consists of three main stages. The first stage is to scan the environment and define the target trajectory of the end-effector, the second stage defines the feasible range of motion of the UAV using inverse kinematics and then designs the UAV's motion plan based on TTC. Lastly, Q-learning is used to modify the joint torques of the manipulator according to the tracked end-effector trajectory providing a robust controller capable of performing in different environments. This method separates the arm controller functionality from the flight controller to maintain the convenience of using arbitrary remote-controlled or autonomous flight controllers with common firmware such as ArduPilot, PX4, etc.

Our proposed method paves the road to utilizing arm-equipped drones for executing complicated tasks in remote, inaccessible, and hazardous environments. We envision that the proposed approach has great potential to impact various fields such as agriculture, industrial, and medical domains. This project also emphasizes the importance of developing a robust control strategy that can plan an optimal trajectory for both the quadrotor and the equipped manipulator while considering their kinematics and dynamics constraints.

\begin{figure}[htbp]
\centerline{\includegraphics[scale=0.38]{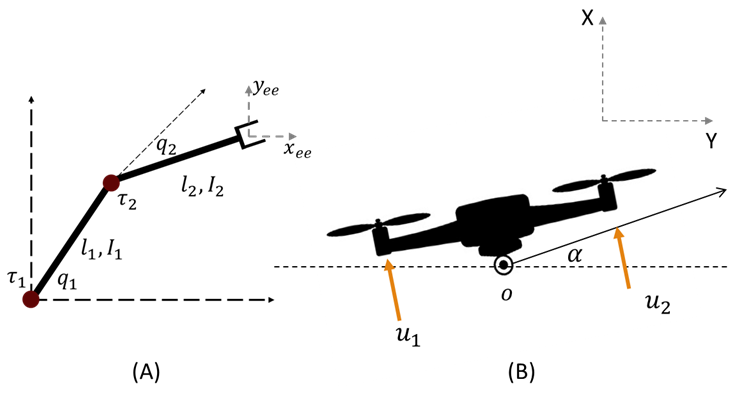}}
\caption{Model Dynamics. (A) represents the two-dimensional body diagram of the manipulator, (B) represents the two-dimensional body diagram of the quadrotor.}
\label{fig:dynamics} 
\vspace{-0.2 in}
\end{figure}

\section{System Model}
\subsection{Model Kinematics}
\label{sec:kinem}
Fig. \ref{fig:dynamics} illustrates the schematics used for calculating and utilizing the model's kinematics to determine the approximate feasible region, which hosts allowable UAV paths that can be used to achieve the end-effector target trajectory. The main idea is to provide the motion planner with the kinematic constraints dictated by the manipulator's joint limits ensuring that the target end-effector trajectory is always reachable from any point on the resultant feasible region\cite{kinem1,kinem2}.

Model inverse kinematics are obtained using Cartesian coordinates \cite{Tedrake} as follows: 
\begin{align}
q_2 &= \cos^{-1}{\frac{x_t^2+y_t^2-l_1^2-l_2^2}{2l_1l_2}}\label{eq1}, \\
q_1 &= \tan^{-1}{\frac{x_t}{y_t}}+\tan^{-1}{\frac{l_1\sin{q_2}},{l_1+l_2\cos{q_2}}}\label{eq2}
\end{align}
where $(x_ee,y_ee)$ the end-effector coordinates and $l_i$ are the arm lengths.
These kinematic equations are essential at different stages of this experiment as they are utilized to determine the kinematic reachability of the aerial manipulation system. Also, they are used to enforce kinematic constraints during the RL training stage, ensuring reasonable joint torques are being fed to the controller at all times.

\subsection{Model Dynamics}
\label{sec:dyn}

A hybrid system consists of two main dynamic parts that can be modeled independently from each other \cite{UAV cntrl}; therefore, we can consider our platform, the UAV along with the equipped manipulator, as a hybrid system. This decision is driven by the assumption that the UAV propellers provide adequate thrust such that the motion of the manipulator does not affect the overall stability of the hybrid system. This assumption is common in robotic motion planning and trajectory optimization \cite{TOWR} and helps simplify the model to serve as a proof of concept. However, it becomes invalid and poses approximation errors when the mass of the manipulator is significant enough to disturb the UAV's motion plan, making it beneficial to adopt a more complex hybrid dynamic model to accommodate such disturbances. We start the kinematics formulation with this assumption for convenience but our model is general and can handle baseline disturbance during the training phase. Indeed, we include some stability analysis in section \ref{sec:stability} showing the effects of the manipulator dynamics of the drone, and the effect of the path deviation on the learned control.


Fig. \ref{fig:dynamics} illustrates the schematics used for calculating the UAV/quadrotor dynamics (A) and the manipulator dynamics (B). For the quadrotor dynamics, the main constraint concerns the angle of attack ($\alpha$), which must be maintained within a certain limit depending on the overall UAV design and its center of mass specifications. The equations of motion governing this behavior can be derived from single rigid body dynamics\cite{Tedrake}, as follows.
\begin{align}
\label{quadrotor1}
m_o \ddot{x_o} &= -(u_1 + u_2)\sin\alpha, \\ 
\label{quadrotor2}
m_o \ddot{y_o} &= (u_1 + u_2)\cos\alpha - mg,  \\ 
\label{quadrotor3}
I_o \ddot\alpha& = r (u_1 - u_2), 
\end{align}

Here, $u_1,u_2$ are the propeller thrust forces, $m_o$ is the mass, $\ddot{x_o},\ddot{y_o}$ are the horizontal and vertical accelerations respectively, and $I$ is the moment of inertia, $\alpha$ is the angle of attack, and $r$ represents the distance between the center of mass to the propeller. 

The dynamic constraints for the manipulator follow a similar rule and are also bound by the kinematic constraints of the system since joint torques are decided by the required joint angles. This is turned into an inverse expression during the training process where the training model uses joint limits to verify the validity of the learned joint torque. The motion equations of the manipulator can be represented in the Lagrangian form, as described in \cite{Raibert}.

This system has two active joints, which means there exist two torque values acting on the two joints. The joint angles ($q1,q2$) are used as input for the derived model. Cartesian coordinates are used to define the position vector of each joint, which is then used to calculate the joint velocity. Joint velocity and joint coordinates are then used to calculate the kinetic and potential energies of the system, respectively. The position vectors are defined as

\begin{align}
&\mathbf{r}_1 = \begin{bmatrix} l_1 \cos q_1 \\ l_1 \sin q_1 \\ 0 \end{bmatrix}, 
&\mathbf{r}_2 = \begin{bmatrix} l_1 \cos q_1 + l_2 \cos q_2 \\ l_1 \sin q_1 + l_2 \sin q_2 \\ 0 \end{bmatrix}.
\end{align}

Position vectors are plugged into the velocity calculations, combining longitudinal and angular velocities that are then used to calculate the kinetic energies of the system, while potential energies are calculated using the position vectors and mechanical properties of the system:

\begin{align}
\mathbf{v}_1 &= \boldsymbol{\omega}_1 \times \mathbf{r}_1   ,\\
\mathbf{v}_2 &= \boldsymbol{\omega}_2 \times \mathbf{r}_2 + {\omega_1} \times \mathbf{r_2} ,\\
T_1 &= \frac{1}{2} m_1 \mathbf{v}_1^2 + \frac{1}{2} \boldsymbol{\omega}_1^T I_1 \boldsymbol{\omega}_1 ,\\
T_2 &= \frac{1}{2} m_2 \mathbf{v}_2^2 ,\\
    V_1 &= m_1 g l_1 \cos q_1 ,\\
    V_2 &= m_2 g l_2 \cos q_2,
\end{align}
where $\omega_1,\omega_2$ are the angular velocities of the manipulator joints,$I_1,I_2$ represent the moment of inertia at each manipulator arm, and $T_1,T_2,V_1,V_2$ are the kinetic and potential energies of the system respectively. The Lagrangian is calculated as the difference between the kinetic and potential energy as $\mathcal{L} = T - V$, which is then plugged into the Lagrangian derivation formula to produce the generalized equation of motion:
\begin{align}
\frac{\mathrm{d}}{\mathrm{d}t} \left( \frac{\partial \mathcal{L}}{\partial \dot{q}} \right) - \frac{\partial \mathcal{L}}{\partial q} = \boldsymbol{\tau},\\
M(q) \ddot{q} + C(q,\dot{q}) \dot{q} = \boldsymbol{\tau}.\label{EOM}
\end{align}


Following are the derived equations for the motion of the two-arm manipulator mounted on the UAV:




\begin{align}
    M(q) &= \begin{bmatrix} I_1 + I_2 + m_2 l_1^2 +
      2m_2 l_1 l_{c2} c_2 & I_2 + m_2 l_1 l_{c2} c_2 \\ I_2 + m_2 l_1 l_{c2} c_2
      & I_2 \end{bmatrix},\label{eq:Hacrobot}\\ 
C(q,\dot{q}) &=
      \begin{bmatrix} 2 m_2 l_1 l_{c2} s_2 \dot{q}_2 & m_2 l_1 l_{c2} s_2
      \dot{q}_2 \\ m_2 l_1 l_{c2} s_2 \dot{q}_1 & 0 \end{bmatrix},\\
\tau(q) &= \begin{bmatrix} m_1 g l_{c1}s_1 + m_2 g (l_1 s_1 +
      l_{c2}s_{1+2}) \\ m_2 g l_{c2} s_{1+2} \end{bmatrix},\label{tau}
\end{align}
where $M(q)$ is the mass matrix, $C(q,\dot{q})$ is the Coriolis and gravity matrix, and $\tau(q)$ is the control torque matrix. Those equations are used in the action constraint enforcement phases in the training model, assuring a physically sound model behavior.

\subsection{Hybrid System Instability} \label{sec:stability}

Unlike Manipulator-equipped Ground Vehicles \cite{UGV} where the system is supported by its ground contacts, Aerial Manipulators have the disadvantage of being susceptible to unstable behavior caused by external conditions as well as the system's own components. In our prototype, the motion of the overhead manipulator creates a challenge in managing the UAV's flight stability due to the shift in the center of mass caused by the manipulator's movement. The shift in the center of mass can lead to changes at the moment 
and thrust acting on the aerial manipulator, impacting its flight stability.

To analyze the effect of the manipulator on the UAV's stability, we consider both static and dynamic moments. Referring back to Fig.\ref{fig:dynamics}, the static moment $M_{static}$ represents the moment caused by the manipulator arms' weight given by:
\begin{align}
    M_{static} = - \frac{L_1}{2}m_1g\sin(q_1-\alpha) +
    \frac{L_2}{2}m_2g\sin(q_1+q_2-\alpha), \label{static}
\end{align}
where $L1,L2$ are the manipulator arm lengths, $\alpha$ is the angle of attack, and $m_1,m_2$ are the manipulator arm masses.

In addition to the static moment, the dynamic moment $M_{dynamic}$ accounts for the effects of the angular acceleration and torques applied by the manipulator's joints:
\begin{align}
    M_{dynamic} = -I_1\ddot{q_1} + I_2(\ddot{q1}+\ddot{q2}), \label{dynamic}
\end{align}
where $I_1,I_2$ are the moments of inertia of the manipulator's arms.
Considering both the static and dynamic moments, the overall moment acting on the aerial manipulator hybrid system is given by:
\begin{align}
    M = M_{static} + M_{dynamic}. \label{moment}
\end{align}

To maintain a stable flight, it is crucial to carefully control and compensate for the effects of these moments. A proper control strategy must be designed to adjust the UAV's attitude and compensate for the moment disturbances caused by the manipulator's motion. 
Thanks to the flexibility of RL methods (e.g., QL, DQN) in modeling intricate relations between actions and outcomes as a black box, our model automatically captures and handles such turbulence if given sufficient real-world data or precise modeling in simulation environments. 
In this paper, we investigate the impact of the manipulator movement in our simulation, under different learning rates. 

\section{Control Models}
\begin{figure}[t]
\centerline{\includegraphics[height=0.35\textheight, width=0.7 \columnwidth]{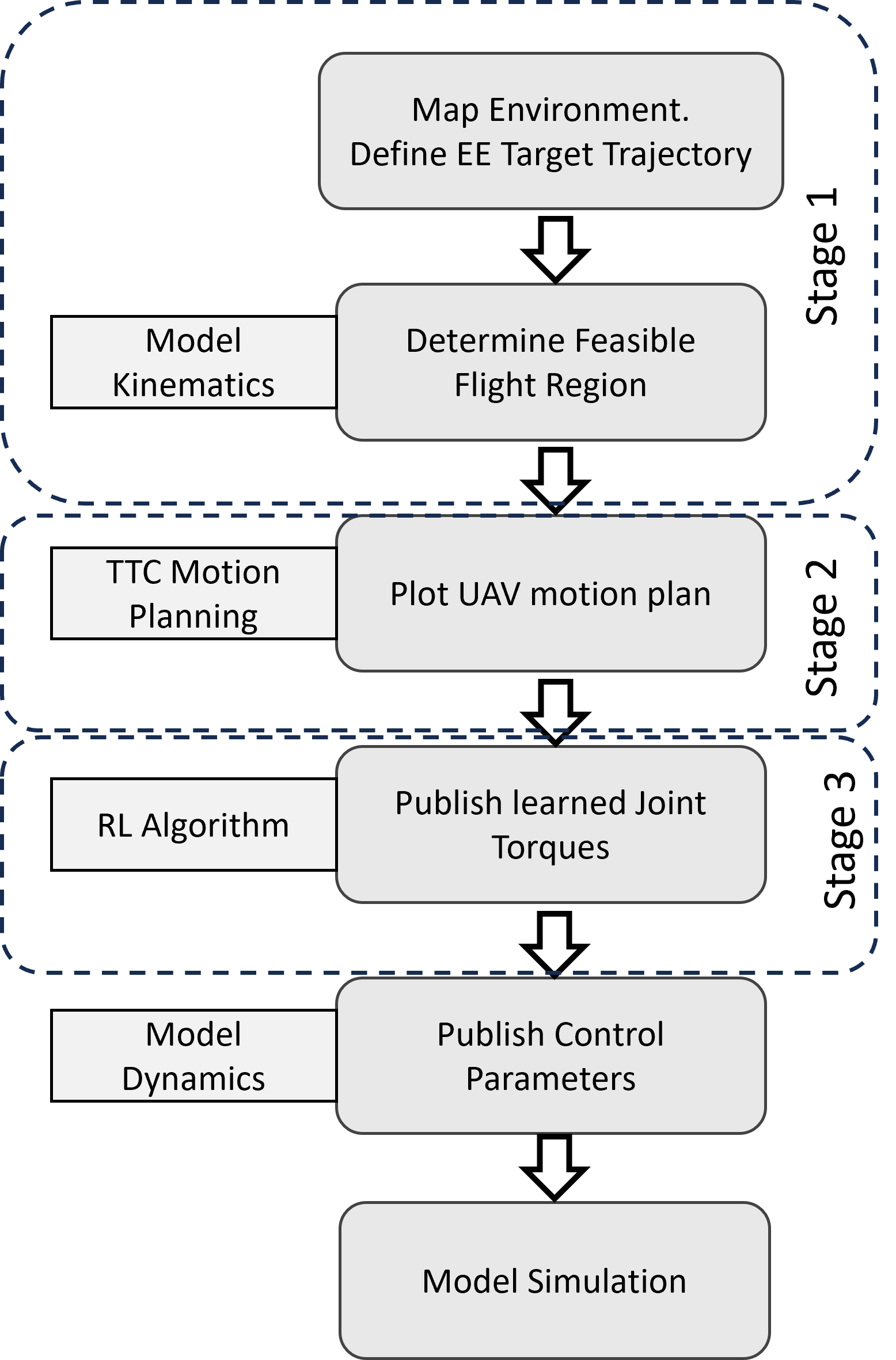}}
\caption{Experimental setup.}
\label{exp} 
\vspace{-0.2in}
\end{figure}

Fig. \ref{exp} depicts the overall experimental setup for our project, consisting of three main stages. In the first stage, 
the environment is mapped into the simulation environment by obtaining the mission-oriented target end-effector trajectory. 
This trajectory is then processed using the system kinematics discussed in Section \ref{sec:kinem} to determine the feasible flight region for the UAV.
In the second stage, a TTC-based motion planning algorithm is utilized to generate a motion plan within the feasible region. This motion plan is subsequently integrated into the RL algorithm along with the target end-effector trajectory, forming the final stage. In this final stage, Q-learning is employed to learn the optimal joint torque sequence for the simulation, enabling the tracking of the end effector trajectory.
 
\subsection{TTC-Based Motion Planning}
Time to Collision (TTC) \cite{TTC,TTCPSO} refers to the time it takes for a moving object (UAV in our case) to collide with an obstacle (considering that all moving objects continue with their current speed) or come to a complete stop without utilizing any braking control system upon completely releasing the throttle. By considering various parameters, this model estimates the time that remains until the UAV collides with an obstacle along its trajectory. In this section, a TTC based motion plan is utilized to provide a motion plan for the UAV within the designated feasible region.
TTC-based planning is known to be highly beneficial in unknown environments with dynamic obstacles.



The fundamental concept behind TTC-based obstacle avoidance revolves around calculating the time it would require for the vehicle to reach the obstacle based on its present velocity and the distance to the obstacle\cite{NH-TTC,TTCMP}. Continuously monitoring and updating these values allows the system to make real-time decisions on steering, braking, or altering the vehicle's speed to prevent a collision.

Implementing TTC-based obstacle avoidance typically involves utilizing sensors such as cameras, LiDAR, or radar to detect and track obstacles in the vehicle's surroundings. These sensors provide information regarding the position, velocity, and size of the obstacles. By combining this data with the vehicle's own speed and trajectory, the system can estimate the TTC.

Once the TTC is determined, the system compares it to predefined safety thresholds or criteria. If the TTC falls below a specific threshold, indicating an imminent collision, the system triggers appropriate evasive actions to avoid the obstacle. 

Once the TTC-based motion planning is completed for the UAV platform, this base trajectory is used as an input for the subsequent RL-based arm controller module that exerts joint torque on the arm joints as discussed in Eq. \eqref{EOM} - Eq.\eqref{tau}, to yield the desired end-effector trajectory.

\subsection{Q-Learning}

In Q-learning algorithms \cite{RL,DRL}, an agent learns to take action in an environment by interacting with it and receiving rewards or punishments for its actions. The goal of the agent is to obtain a policy that maximizes the total reward it receives over time. This can be mathematically represented by the following optimization problem:

\begin{align}
arg\max_{\pi} \mathbb{E} \sum_{t=0}^ T  \gamma^t R_t | \pi,
\end{align}
where 
$R_t|\pi=R_t(s_t,a_t=\pi(s_t),s_{t+1})$ is the reward received at time $t$ starting from an initial state $s_0$ and following policy $\pi: \mathcal{S} \mapsto \mathcal{A}$ that maps states state $s_t \in \mathcal{S}$ at time point $t$ to action $a_t \in \mathcal{A}$. The expectation is taken over all possible states noting that $s_t \xrightarrow{a_t} s_{t+1}$ transitions are probabilistic in general for Markov Decision Processes (MDPs). The horizon $T$ can be finite or infinite. The discount factor $0 < \gamma \leq 1 $ is used to promote collecting awards faster and also to have a bounded total reward for an infinite horizon. In our case, states are the discretized values of the current position of the drone and its orientation (base position) as well as the current joint angles, and actions are the applied torque to the arm manipulator motors to adjust angles $q_1,q_2$, as detailed in the sequel. 

To solve this optimization problem, the agent maintains a value function $Q(s, a)$, which represents the expected total reward the agent will receive if it takes action $a$ in state $s$ and follows the optimal policy thereafter. The value function can be updated using the following Bellman equation:

\begin{align*}
Q(s,a) \leftarrow Q(s,a) + \alpha [R + \gamma \max_{a'} Q(s',a') - Q(s,a)],
\end{align*}
where $s'$ is the next state of the system, $a'$ is the action taken in that state, $\alpha$ is the learning rate. 

To solve the tracking problem for the manipulator tip using Q-learning, we would need to define the state space, action space, reward function, and transition model. 


The state space in this experiment can be defined as the UAV pose data and current joint states $[q_1,q_2,t_o]$, and the action space for the training model, in this case, is the joint torque commands $[\tau_1,\tau_2]$. The reward function is calculated based on the distance between the obtained 
end-effector trajectory and the desired target trajectory.
The transition model describes how the state of the system changes based on the current state and the action taken by the agent.

To define the transition model for our system, the equations of motion of the manipulator are taken into consideration. 
Specifically, if the current state of the system is represented by the vector $\mathbf{s} = \begin{bmatrix} q_1 & q_2 & pose_{UAV} \end{bmatrix}^T$, and the action taken by the agent is represented by the vector $\mathbf{a} = \begin{bmatrix} \tau_1 & \tau_2 \end{bmatrix}^T$, then the transition model can be expressed as a function $\mathbf{s}' = f(\mathbf{s},\mathbf{a})$ that takes the current state and action as inputs and returns the next state of the system.$\mathbf{s}' = f(\mathbf{s},\mathbf{a})$ is described in the model dynamic in section \ref{sec:dyn}.

\section{Simulation Results}

\begin{figure}[t]
\centerline{
\includegraphics[height=3.2in,width = 3.1 in]{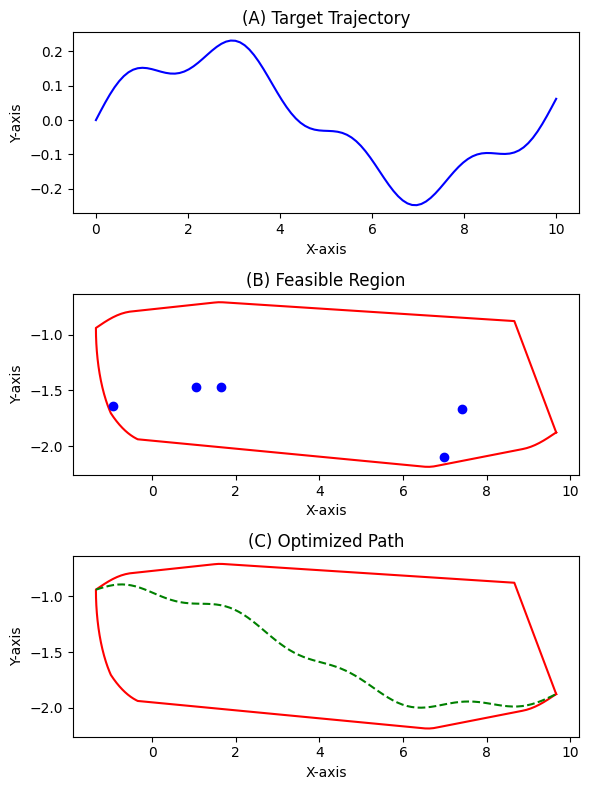}
}
\caption{Top: Target end-effector trajectory, middle: Generated feasible region with obstacles, bottom: TTC-based UAV trajectory plan(dashed-green).}
.
\label{TTC} 
\vspace{-.2 in}
\end{figure}

The overall experimental setup designed for this project is shown in Fig.\ref{exp}, including three stages of environment scan, TTC-based motion planning, and Q-learning training for end-effector trajectory. The first step is to acquire the target end-effector trajectory in the simulation environment as shown in Fig.\ref{TTC}, which is then plugged into the system kinematic calculations to produce the feasible region for the UAV flight path; next, a TTC-model is used to plot the best motion plan for the UAV within that region such that the end-effector trajectory is reachable at all times. 

\begin{figure}[t]
\centerline{\includegraphics[scale=0.45]{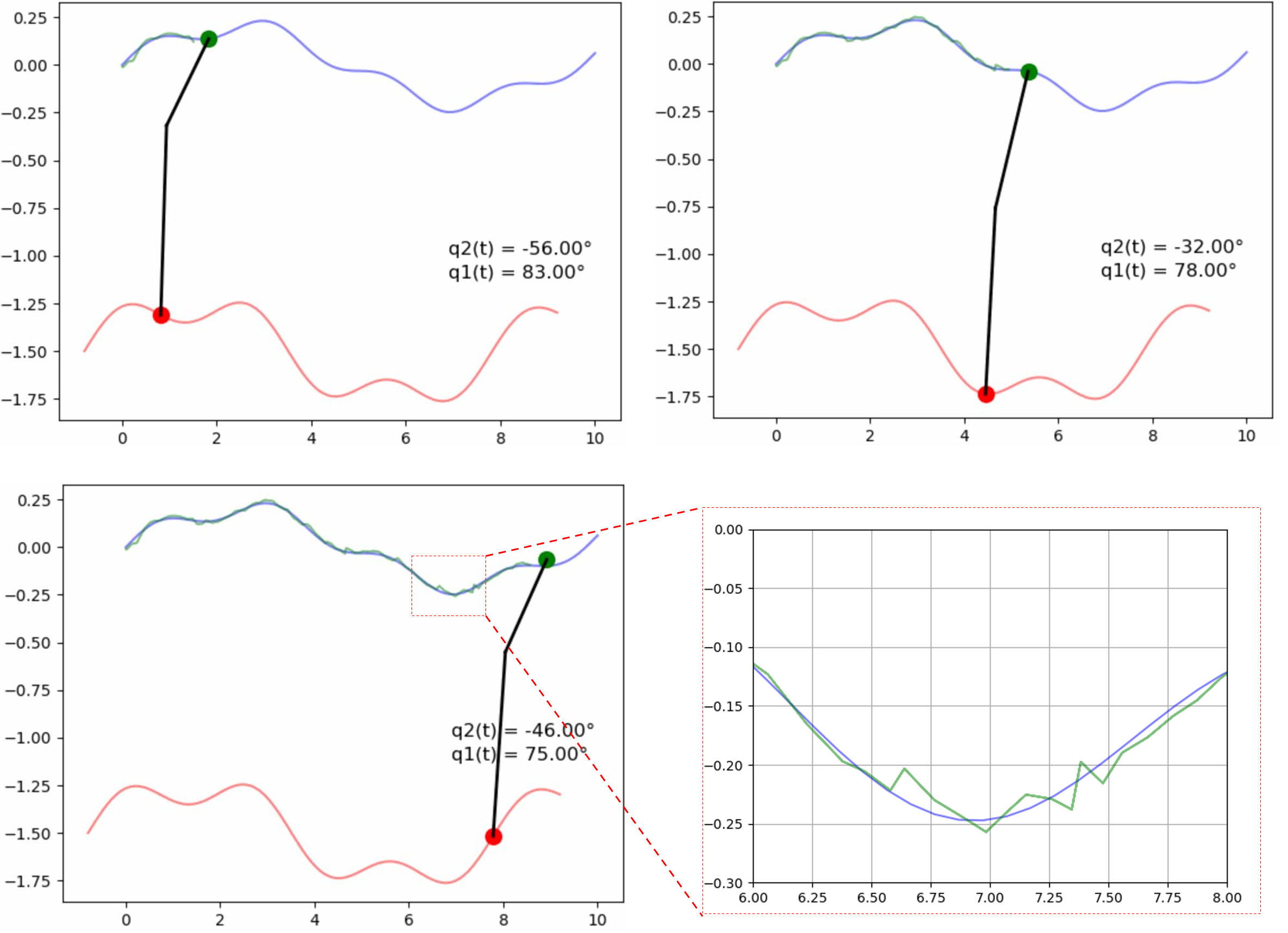}}
\caption{Q-learning results. The manipulator is able to track its end-effector target trajectory with an overall accuracy of \%92.}
\label{QL2} 
\end{figure}

The last phase involved executing a reinforcement learning algorithm over 15,000 episodes, utilizing a learning rate of 0.1. The outcome exhibited an average reward of 7.96 (the maximum reward is 10) and a Root mean Squared Error (RMSE) value of 0.08. As shown if Fig.\ref{QL2}, the algorithm effectively regulates the joint coordinates by employing joint torque commands to follow the desired end-effector trajectory. The performance of the algorithm in facilitating appropriate state transitions during the simulation process was commendable. However, it should be noted that the learned trajectory appears discontinuous due to the discretization of the model's state space and action space. 

An analysis of the effect of learning parameters and discretization samples on the overall performance of the learning agent is presented in Table. \ref{tab:rmse}. It can be seen that reducing the learning rate from $\alpha=0.1$ to $\alpha=0.001$ negatively affects the performance of the RL agent. Also, a discount factor of 0.9 shows the highest performance since using an extremely low discount factor (like $\gamma=0.2$) undermines the optimality of the selected solution. 
As expected increasing the number of samples improves the performance of the algorithm at the cost of longer training time.

\begin{table}[t]
  \centering
  \caption{Impact of RL parameters on the simulation, learning algorithm ran for 15,000 episodes}
  \label{tab:rmse}
  \setlength{\tabcolsep}{8pt}
  \renewcommand{\arraystretch}{1.2} 
  \begin{tabular}{|c|c|c|c|}
    \cline{3-4}
    \multicolumn{2}{c|}{} & \textbf{RMSE} & \textbf{Avg Reward} \\
    \hline
    \multirow{3}{*}{Learning rate} & 0.1 & 0.08 & 7.93 \\
    \cline{2-4}
                           & 0.01 & 0.093& 5.54 \\
    \cline{2-4}
                           & 0.001 & 0.11& -1.89 \\
    \hline
    \multirow{3}{*}{Discount factor} & 0.9 & 0.08& 7.93 \\
    \cline{2-4}
                           & 0.5 & 0.085&3.63 \\
    \cline{2-4}
                           & 0.2 & 0.089& 1.92 \\
    \hline
    \multirow{3}{*}{Number of samples} & +\%25 &0.098& 5.77 \\
    \cline{2-4}
                           & -\%25 & 0.16& -2.67 \\
    \hline
  \end{tabular}
\end{table}

\begin{figure}[t]
\centerline{\includegraphics[height=3.2in,width = 3.2 in]{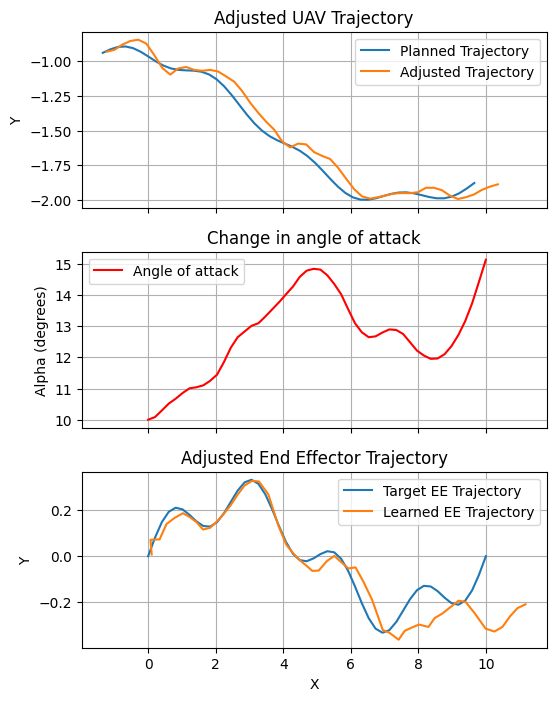}}
\caption{Effects of the manipulator motion}
\label{result2} 
\vspace{-0.2 in}
\end{figure}

In order to examine how dynamic instability influences the system's behavior, we abandoned the simplifying assumption that the manipulator arm motion has a negligible effect on the UAV's base trajectory explained in \ref{sec:dyn}. To this end, we amplify the masses of the manipulator's arm, leading to the activation of Eq.\eqref{moment}. By altering the moment at the center, we intentionally destabilized the UAV, resulting in an immediate deviation from its intended trajectory. Moreover, this destabilization affected the angle of attack, increasing the risk of surpassing its limitations.

Fig. \ref{result2} illustrates the influence of the manipulator's movement on the trajectory and angle of attack of the UAV. In order to mitigate these effects, a basic reactive model was utilized to regulate the thrust forces, as described by Eq.\eqref{quadrotor3}, with the objective of maintaining the intended trajectory. However, due to response delays, the UAV faces difficulty in fully restoring its planned motion. This challenge arises from the active motion of the manipulator, which generates a non-constant moment. Furthermore, the deviation in the UAV's motion disrupts the acquired joint torques, further deteriorating the overall system performance. Although the RL model was successful in managing disruptions caused by deviations in the planned motion, it encountered more difficulty in overcoming disturbances at points where there were significant changes in the angle of attack. This difficulty stems from the relationship between joint angles and the angle of attack in Cartesian space.


\section{Conclusion}
A three-stage RL-based solution is developed to carry out complicated tasks on the fly using drones with overhead manipulators. Our solution includes three stages, i) the acquisition and mapping of the target end-effector trajectory to generate a feasible UAV flight path using kinematic calculations, ii) employing a TTC-based path planning for the drone center of mass within the feasible region, ensuring continuous reachability of the end-effector trajectory, and iii) using Q-learning algorithm to control the torque acting on the manipulator's joint coordinates to track the target trajectory. 
We showed that the proposed approach in the simulation environment achieves a promising performance of 90\% plus accuracy in following the mission-oriented trajectory. This is achieved by the separation of drone path planning and mounted actuator control to keep compatibility with arbitrary path planners and flight control firmware, such as ArduPilot, PX4, and more. 

Additionally, we investigated the instability of the drone flight path caused by the motion of the overhead manipulator which can alter the center of mass and the angle of attack. We showed that RL-based algorithms are capable of handling mild instabilities without a significant deviation from the desired target trajectory given that the simulation environment incorporates such static and dynamic instabilities or sufficient training data points are provided for such situations. This work underscores the need for developing more robust RL algorithms and employing techniques such as model-predictive control, closed-loop control, and adaptive control to improve the overall performance and reliability of the actuator drone operation under more severe disturbances.

\vspace{12pt}

\end{document}